# A Generalist Foundation Model for Total-body PET/CT Enables Diagnostic Reporting and System-wide Metabolic Profiling

Wei Chen[1†], Liang Wu[1†], Shuyi Lu[1†], Yuanyuan Sun[1†], Wenkai Bi[2], Zilong Yuan[3], Yaoyao He[3], Feng Wang[4], Junchi Ma[1], Shuyong Liu[4*], Zhaoping Cheng[5*], Xiaoyan Hu[6*], Jianfeng Qiu[5*]

[1] Imaging Supercomputing Centre, Shandong First Medical University & Shandong Academy of Medical Sciences, Jinan 250000, China

[2] Department of Nuclear Medicine, Shandong Provincial Hospital, Shandong First Medical University, Jinan 250000, China

[3] Department of Radiology, Hubei Cancer Hospital, Tongji Medical College, Huazhong University of Science and Technology, Wuhan 430000, China

[4] Department of Nuclear Medicine, The Second Affiliated Hospital of Shandong First Medical University, Taian 271000, China

[5] Department of Nuclear Medicine, The first Affiliated Hospital of Shandong First Medical University, Jinan 250000, China

[6] Department of Nuclear Medicine, Hubei Cancer Hospital, Tongji Medical College, Huazhong University of Science and Technology, Wuhan 430000, China

*Corresponding author(s).

†These authors contributed equally to this work.

**Abstract**

Total-body positron emission tomography/computed tomography (PET/CT) extends molecular imaging from organ-level snapshots to system-wide observation of human physiology and disease. Yet the very features that make this modality transformative—heterogeneous anatomical and metabolic signals, ultra-long axial coverage, and highly structured radiological semantics—also expose fundamental limitations of current medical AI models, which are largely optimized for single-modality inputs, localized fields of view, and coarse image–report alignment. Here we introduce SDF-HOLO (Systemic Dual-stream Fusion Holo Model), a multimodal generalist foundation model designed for holistic analysis of total-body PET/CT and trained on a large-scale cohort of more than 10,000 patients. SDF-HOLO decouples anatomical (CT) and metabolic (PET) representation learning through a dual-stream encoder and enables bidirectional synergy via a Cross-Modal Interaction Module, allowing anatomical priors to refine PET signal aggregation while metabolic saliency guides subtle morphological reasoning. To capture body-wide dependencies across the ~2-meter axial field of view, SDF-HOLO incorporates hierarchical context modeling that integrates efficient local windows with global attention. To bridge the semantic gap between voxels and clinical language, we further introduce anatomical segmentation masks as explicit semantic anchors, establishing a voxel–mask–text tripartite alignment during pre-training for fine-grained spatial–semantic correspondence. Across automated tumor segmentation, low-dose lesion detection, and multi-lingual diagnostic report generation, SDF-HOLO consistently outperforms strong task-specific and clinical-reference baselines while reducing localization errors and hallucinated findings.

Beyond focal interpretation, the model enables system-wide metabolic profiling and reveals tumor-associated fingerprints of inter-organ metabolic network interactions. Together, SDF-HOLO provides a scalable computational foundation for total-body PET/CT–driven diagnostics and system-level precision oncology.

1. Introduction

Medical artificial intelligence is undergoing a rapid transition from task-specific decision support toward generalist foundation models that can transfer across tasks, organs, and institutions. In the medical imaging domain, recent work has begun to formalize how vision foundation models can be adapted to heterogeneous multi-center settings and multiple downstream tasks, highlighting the emerging paradigm of scalable, reusable representations [1-3]. In parallel, comprehensive surveys of multimodal learning have emphasized that deep fusion across imaging modalities and clinical narratives is becoming a key route to improved robustness and generalization in real-world workflows [2].

In this context, automated radiology reporting has become a prominent testbed for medical vision–language modeling, yet it remains challenging due to the need for faithful evidence grounding, reduced hallucination, and clinically meaningful evaluation beyond generic NLG metrics [4, 5]. Recent studies have proposed multimodal LLM frameworks for volumetric (3D) CT report generation and clinical instruction tuning, underscoring both the promise and unresolved bottlenecks of 3D radiology-language alignment [6]. Meanwhile, segmentation-assisted report generation and the construction of paired CT–mask–report datasets further suggest that explicit spatial supervision can substantially improve report fidelity and lesion attribution [7].

Alongside these algorithmic advances, imaging hardware has entered a similarly transformative era. Total-body PET/CT extends molecular imaging from regional, multi-bed acquisitions to near whole-body, single-pass capture, enabled by long axial fields of view approaching ~2 meters in modern systems [8]. This capability increases sensitivity, supports ultra-low-dose imaging protocols, and allows synchronized observation of cross-organ physiology and disease processes [9]. Practical performance characterization of long-AFOV clinical systems (e.g., Biograph Vision Quadra) has further clarified how acquisition modes (such as continuous bed motion) and ultrahigh sensitivity settings affect quantitative metrics and image quality [10].

Despite these opportunities, total-body PET/CT introduces data characteristics that expose fundamental limitations of mainstream medical AI. First, modality heterogeneity often leads to representation imbalance: CT contains dense anatomical textures, whereas PET signals are sparse and functionally salient, and naive fusion can suppress metabolically meaningful cues. PET/CT-specific foundation-model pretraining efforts have begun to address this issue via separate encoders and cross-modal interactions during representation learning [11]. Second, whole-body coverage and heterogeneous protocols amplify domain shift; for example, large-scale PET/CT lesion segmentation benchmarks have shown that out-of-domain generalization remains a major barrier in clinically realistic settings [12]. Third, ultra-low-dose and CT-free workflows create additional challenges for reliable quantification and organ-level understanding; recent multicenter evidence supports the feasibility of PET-only multi-organ segmentation across tracers and acquisition conditions, and reader-validated low-count PET denoising for dose/time reduction [13, 14].

To address these challenges in a unified manner, we present SDF-HOLO (Systemic Dual-stream Fusion Holo Model), a multimodal generalist foundation model designed specifically for total-body PET/CT and aimed at enabling a shift from focal lesion analysis to system-level understanding. SDF-HOLO is built on three core ideas: (i) an anatomical–metabolic decoupled dual-stream encoder with bidirectional cross-modal interaction; (ii) hierarchical whole-body context modeling for ultra-long axial reasoning; and (iii) mask-guided fine-grained semantic anchoring that enforces voxel–mask–text tripartite alignment to strengthen spatial–semantic correspondence and mitigate hallucinations.

## 2. Related Work

Generalist foundation models are reshaping medical image analysis, moving the field from narrowly optimized, task-specific pipelines toward reusable representations that can be adapted across organs, scanners, and clinical endpoints. Total-body PET/CT magnifies both the opportunity and the difficulty of this transition: it couples high-resolution structural context (CT) with sparse, high-dynamic-range functional signals (PET) across an ultra-long axial field-of-view, and it must support heterogeneous downstream objectives ranging from voxel-level tumor delineation and low-dose lesion detection to evidence-grounded report generation and system-level metabolic profiling. Below, we review related progress in medical imaging foundation models, PET/CT-specific multimodal learning, whole-body PET/CT benchmarks, radiology vision-language modeling, and organ-level metabolic interaction analysis.

### 2.1 Foundation models for medical imaging

Self-supervised pre-training has become a dominant strategy for learning transferable medical imaging representations, particularly when dense labels are scarce or expensive. Early work adapted contrastive learning and masked image modeling to medical data, and more recent efforts have scaled these ideas into "foundation" or "generalist" models spanning multiple modalities and tasks [15-17]. Radiology-focused foundation models further integrate paired text (reports, captions, instructions) with images to unify classification, retrieval, reasoning, and generation within a shared representation space, improving transfer and enabling more flexible clinical interfaces [18].

A key technical distinction is between 2D-centric models and native 3D approaches. While 2D pre-training can leverage very large image collections, volumetric reasoning is essential for CT and PET/CT, where lesion extent, anatomical continuity, and cross-slice context affect clinical decisions.

Recent 3D vision-language foundation models explicitly handle volumetric inputs and demonstrate improved transfer to 3D downstream tasks. Complementary lines of work pursue universal or modality-adaptive segmentation models trained on large-scale whole-body annotations or prompts, providing standardized anatomical scaffolds that can be reused across applications and modalities [19]. Together, these trends motivate foundation models that are not only large, but also architecturally aligned with volumetric, anatomically structured clinical data.

2.2 Multimodal learning for PET/CT

PET/CT differs from many multimodal learning settings because the two modalities encode complementary physics: CT provides detailed anatomy, whereas PET measures tracer uptake and thus physiology and disease metabolism. Conventional fusion strategies include early fusion (channel stacking), late fusion (independent encoders with feature concatenation), and intermediate fusion (learned interaction such as cross-attention or gated modulation). However, naive fusion often biases optimization toward the higher-frequency CT statistics, which can suppress PET saliency and degrade sensitivity to weak uptake patterns (e.g., small or low-contrast lesions), especially under low-count or ultra-low-dose protocols.

PET/CT-specific self-supervised learning further supports the value of decoupled representation learning. Cross-modal masked autoencoding has been proposed with separate PET and CT encoders and cross-attention decoders to encourage synergistic reconstruction and alignment between anatomy and function during pre-training [20]. In parallel, deep learning methods for count reduction, denoising, and accelerated acquisition aim to preserve diagnostic quality under low-count or ultrafast protocols, making robust modeling of sparse functional signals central to clinical deployment [21, 22]. These developments suggest that effective PET/CT models should preserve modality-specific inductive biases while enabling targeted, bidirectional information exchange.

2.3 Total-body PET/CT and whole-body lesion analysis

Long-axial-field-of-view PET/CT systems, including total-body scanners with near whole-body coverage in a single pass, enable high sensitivity, ultra-low-dose imaging, and synchronized observation of cross-organ physiology and disease processes [23-25]. These hardware advances broaden applications from lesion-centric staging toward system-wide phenotyping, but they also introduce heterogeneous data characteristics that stress current AI systems, including scanner and protocol variability, reconstruction differences, uptake-time variation, and strong physiologic tracer accumulation that can mimic or obscure pathology.

Whole-body tumor segmentation has become a prominent benchmark for PET/CT, in part due to the AutoPET challenges, which emphasize cross-center generalization and clinically realistic hard cases such as small lesions, low uptake, and ambiguous boundaries [26]. Across competing methods, a recurring trade-off emerges between overlap-based metrics (e.g., Dice) and clinically meaningful error profiles such as missed tumor burden (false-negative volume) or spurious uptake-driven errors

(false-positive volume), motivating evaluation protocols that explicitly quantify both under- and over-segmentation. Beyond challenge settings, pan-cancer whole-body lesion segmentation efforts highlight the need for models that remain robust across disease types and anatomical locations, often leveraging both PET intensity cues and CT anatomy.

Multi-organ and multi-structure segmentation is another enabling direction because it provides anatomical context and supports standardized quantification. Large-scale whole-body segmentation resources and universal models can act as anatomical scaffolds, enabling region-aware processing, organ-level measurements, report structuring, and population-level phenotyping. In PET-centric workflows, multicenter evidence suggests that robust multi-organ segmentation is feasible even in ultra-low-dose total-body PET and across tracers, and that deep learning denoising can further improve image quality for dose or time reduction. Together, these studies indicate that the next generation of PET/CT intelligence should jointly model lesions and anatomy under realistic domain shift.

2.4 Vision-language modeling for radiology reporting

Automated report generation is a stringent test of multimodal medical AI because it requires linguistic coherence while remaining faithful to imaging evidence. Systematic reviews document rapid progress but persistent failure modes, including hallucinated findings, incorrect anatomical localization, and omission of clinically salient observations [27]. Recent multimodal large language model (MLLM) frameworks for volumetric CT reporting and clinical instruction tuning demonstrate improved fluency and task generalization, but also underscore that volumetric grounding and safety-oriented evaluation remain major bottlenecks for clinical adoption [28, 29].

To mitigate hallucination and improve localization, a growing body of work augments vision-language training with explicit spatial supervision. Segmentation-assisted report generation and paired image-mask-text resources suggest that anatomical masks can function as semantic anchors that constrain where a textual entity may be supported in the image, improving grounding and reducing unsupported statements [30]. This perspective aligns with broader radiology foundation model efforts that integrate images and rich text supervision to support diverse generative tasks within a single architecture. For total-body PET/CT, where a single study spans many anatomical regions and multiple potential disease sites, fine-grained spatial-semantic correspondence is especially critical for faithful reporting.

2.5 System-level metabolic profiling and interaction modeling

Total-body PET/CT creates a unique opportunity to quantify physiology at the scale of organs and systems. Classical approaches rely on interpretable scalar summaries (e.g., SUVmean/SUVmax) and radiomics features, but representation learning enables richer embeddings that may capture metabolic phenotypes beyond a single intensity value. Building on these representations, investigators have begun to study organ-organ coupling patterns using correlation, covariance, and graph-based analyses, aiming to summarize coordinated physiological variation, identify disease-associated systemic fingerprints, and provide compact descriptors for downstream prediction.

Whole-body universal modeling has been used to reveal brain-body metabolic correlations and systemic associations that are difficult to study with region-limited imaging, suggesting that unified models can facilitate integrative brain-body research and cross-organ mechanism discovery. In oncology, system-level signatures are being explored as potential biomarkers reflecting tumor-host interactions, treatment effects, or comorbidity patterns. Nevertheless, metabolic association analyses are sensitive to confounders such as tracer kinetics, patient preparation, uptake time, and reconstruction settings, motivating reference atlases and confounder-aware validation strategies when translating network-level findings into clinically actionable endpoints.

3. Methods

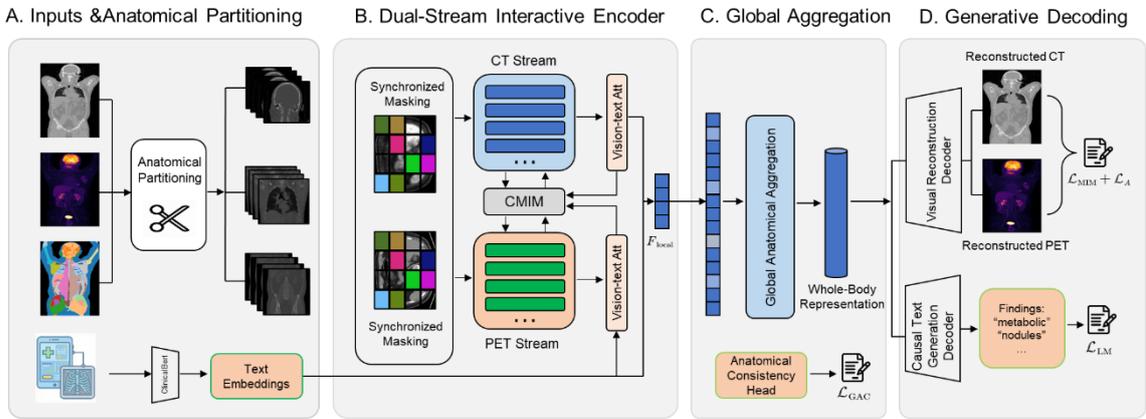

Fig. 1 Architecture of the SDF-Holo whole-body pre-training framework. (a) Inputs & Anatomical Partitioning: Whole-body CT, PET, and Segmentation Masks are processed alongside Radiology Reports. (b) Dual-Stream Interactive Encoder. (c) Global Anatomical Aggregation. (d) Dual-Branch Generative Decoding & Objectives.

3.1 Whole-body pre-training framework

A total-body PET/CT scan spans nearly 2 meters along the axial direction. Directly feeding the full 3D volume into a Transformer is computationally prohibitive due to quadratic attention complexity. Moreover, naive fusion with a single visual encoder often causes functional PET signals to be dominated by high-frequency CT textures. To address these challenges, SDF-HOLO adopts a three-stage paradigm: Partition-awareness → Dual-stream interaction → Global aggregation, enabling efficient local representation learning while preserving systemic context.

As illustrated in Fig. 1, the framework orchestrates this paradigm by processing multi-modal inputs through anatomical partitioning, extracting features via a dual-stream interactive encoder, and synthesizing a systemic representation using global aggregation to support dual-branch generative tasks.

Each study consists of a whole-body CT volume $X_{\text{CT}}$, a co-registered PET volume $X_{\text{PET}}$, and the paired radiology report $T$. In addition, we construct an anatomical segmentation mask volume $S$ as an explicit semantic anchor.

We generate anatomical masks by merging two automated whole-body CT segmentation systems: TotalSegmentator and CADS (CADS: A Comprehensive Anatomical Dataset and Segmentation for Whole-Body Anatomy in Computed Tomography), both applied to $X_{\text{CT}}$. We harmonize their label ontologies into a unified set of 180 non-overlapping anatomical regions $\mathcal{C} = \{1, ..., 180\}$ via:

1) Label mapping: map each source label to a canonical region ID in $\mathcal{C}$ using a manually curated lookup table (synonyms and laterality are normalized). 2) Priority-based fusion: for voxels labeled by both sources with different classes, we prioritize CADS when the target class exists in CADS; TotalSegmentator is used to fill the remaining classes not covered (or missing) in CADS. 3) Topology cleanup: remove small disconnected islands (fewer than $\delta$ voxels), fill holes, and enforce a single-label constraint per voxel.

The resulting mask volume is $S \in \{0, 1, ..., 180\}^{H \times W \times D}$, where 0 denotes background. This mask is used during pre-training to provide geometry-aware semantic anchoring; it is not required at inference time for downstream tasks.

We split each whole-body scan into $R$ anatomically meaningful sub-regions $\{X^{(r)}\}_{r=1}^{R}$ to control token length and encourage region-specific specialization with shared weights. In our implementation, we use $R = 6$: head/neck, thorax, upper abdomen, pelvis, upper limbs, and lower limbs. Partition boundaries are computed using robust axial landmarks derived from $S$ (e.g., superior/inferior extents of lungs, liver, bladder, femurs); if a landmark is unavailable, we fall back to fixed z-percentile boundaries.

3.2 Dual-stream interactive encoder and global anatomical aggregation

For each region $r$, CT and PET are tokenized into 3D patches of size $16 \times 16 \times 16$ voxels and linearly projected into embeddings:

$$Z_{m,r}^{(0)} = [E_{\text{pos}} + \text{Proj}(x_{m,r}^i)]_{i \in \mathcal{V}}, \quad m \in \{\text{CT}, \text{PET}\}.$$

We use synchronized high-ratio masking (masking ratio $p=0.90$) so that CT and PET share the same visible patch index set $\mathcal{V}$, which stabilizes cross-modal learning. Each modality is processed by an EvaMAE-style Transformer encoder stream. To achieve inter-modal (CT↔PET) feature synergy, we insert a Cross-Modal Interaction Module (CMIM) at selected layers (e.g., at the 1/3 and 2/3 depth of the encoder).

CMIM performs bidirectional cross-attention:

$$\text{Attn}_{\text{CT} \leftarrow \text{PET}} = \text{Softmax}\left(\frac{Q_{\text{CT}} K_{\text{PET}}^{\top}}{\sqrt{d_k}}\right) V_{\text{PET}}, \quad \text{Attn}_{\text{PET} \leftarrow \text{CT}} = \text{Softmax}\left(\frac{Q_{\text{PET}} K_{\text{CT}}^{\top}}{\sqrt{d_k}}\right) V_{\text{CT}}.$$

Residual connections and a learnable gate produce the fused regional visual feature $F_{\text{vis}}^{(r)}$, allowing CT structure to guide PET aggregation while PET saliency highlights subtle CT abnormalities.

A key limitation of conventional vision–language pre-training is that it aligns whole volumes with whole reports, leaving spatial–semantic correspondence underconstrained. We therefore use $S$ as an explicit semantic anchor and enforce a voxel–mask–text tripartite alignment.

For each anatomical class $c \in \mathcal{C}$, we compute a pooled embedding from the fused visual tokens inside that class:

$$v_c^{(r)} = \text{Pool}\left(\{F_{\text{vis},i}^{(r)} \mid S_i = c\}\right)$$

In parallel, we encode the report text with a Transformer text encoder to obtain token embeddings:

$$E_{\text{text}} = \text{TextEnc}(T) \in \mathbb{R}^{L_t \times d}.$$

We construct text anchors $t_c$ by pooling the embeddings of anatomical mention spans corresponding to class $c$ (via a curated lexicon covering synonyms and laterality; for multilingual reports, the lexicon includes bilingual terms).

We optimize an InfoNCE objective that brings matched $(v_c, t_c)$ pairs closer while pushing unmatched pairs apart within a minibatch:

$$-\log \frac{\exp(\langle v_c, t_c \rangle / \tau)}{\sum_{c'} \exp(\langle v_c, t_{c'} \rangle / \tau)}$$

This anchor loss explicitly links localized voxel evidence to anatomical geometry and language, reducing localization drift and improving grounding for generation.

To provide report-level global context for regional representation, we optionally inject $E_{\text{text}}$ into the regional visual stream through cross-attention:

$$F_{\text{local}}^{(r)} = \text{MultiHead}(Q = F_{\text{vis}}^{(r)}, K = E_{\text{text}}, V = E_{\text{text}}) + F_{\text{vis}}^{(r)}.$$

This yields semantically enhanced regional features $F_{\text{local}}^{(r)}$.

We reassemble regional features in anatomical order and model long-range systemic dependencies:

$$S_{\text{global}} = \text{Concat}(F_{\text{local}}^{(1)}, \ldots, F_{\text{local}}^{(R)}) + E_{\text{anatomy}},$$

where $E_{\text{anatomy}}$ encodes region identity. A lightweight Transformer (GAA) produces the whole-body representation:

$$F_{\text{whole}} = \text{GAA}(S_{\text{global}}).$$

3.3 Unified generative decoders and objective functions

A lightweight Transformer decoder reconstructs masked CT and PET patches. For region $r$, the decoder restores full token order by inserting learnable mask tokens:

$$H_{\text{dec}}^{(r)} = \text{RestoreOrder}([F_{\text{vis}}^{(r)}, M_{\text{mask}}]) + E_{\text{pos}}^{\text{dec}}.$$

Two modality-specific linear heads output $\hat{x}_{\text{CT}}^{(r)}$ and $\hat{x}_{\text{PET}}^{(r)}$. The MIM loss is:

$$\mathcal{L}_{\text{MIM}} = \sum_{r=1}^{R} \sum_{i \in \mathcal{M}} \|\hat{x}_{\text{CT},i}^{(r)} - x_{\text{CT},i}^{(r)}\|_2^2 + \omega_{\text{PET}} \|\hat{x}_{\text{PET},i}^{(r)} - x_{\text{PET},i}^{(r)}\|_2^2),$$

where $\omega_{\text{PET}} = 3$ compensates for PET sparsity and dynamic range.

A causal Transformer language decoder generates the report autoregressively conditioned on $F_{\text{whole}}$:

$$\mathcal{L}_{\text{LM}} = -\sum_{t=1}^{L} \log P_\theta(w_t \mid w_{<t}, F_{\text{whole}}).$$

To preserve spatial awareness under partitioning and weight sharing, we add a region classification auxiliary task:

$$\mathcal{L}_{\text{GAC}} = -\sum_{r=1}^{R} \sum_{c=1}^{R} y_{r,c} \log \hat{y}_{r,c}.$$

The final objective combines reconstruction, generation, and semantic anchoring:

$$\mathcal{L}_{\text{total}} = \lambda_1 \mathcal{L}_{\text{MIM}} + \lambda_2 \mathcal{L}_{\text{LM}} + \lambda_3 \mathcal{L}_{\text{GAC}} + \lambda_4 \mathcal{L}_{\text{A}}.$$

Here we set $(\lambda_1, \lambda_2, \lambda_3, \lambda_4) = (1.0, 1.0, 0.1, 0.5)$.

## 4 Results

### 4.1 Dataset Construction and Preprocessing

Pre-training SDF-HOLO requires a triplet dataset comprising co-registered PET/CT volumes, paired nuclear-medicine reports, and anatomical segmentation masks. We assembled a multi-center dataset by retrospectively collecting matched PET/CT–report pairs from nuclear medicine departments at four tertiary hospitals in China, yielding 10,460 studies in total. Each study includes a CT volume, a PET volume acquired in the same examination, and the corresponding clinical report for the individual patient. Data were contributed by 4 centers belong to Shandong First Medical University and our cooperative institution. To our knowledge, this cohort represents one of the largest PET/CT–report resources primarily derived from an Asian population to date.

### 4.2 Downstream Fine-tuning on AutoPET and Benchmark Results

To assess the transferability and cross-domain robustness of SDF-HOLO representations on a standardized public benchmark, we fine-tune the model under the official AutoPET Challenge protocol and evaluate it on the Imu and UKT FDG test subsets. Compared with single-center cohorts, AutoPET aggregates multi-institution, multi-scanner PET/CT studies with heterogeneous reconstruction settings, closely reflecting domain shift in clinical deployment. The benchmark further contains many small, low-uptake, or poorly delineated lesions, providing a stringent test for sensitivity to weak metabolic signals and ambiguous boundaries.

We use sliding-window inference to cover full volumes, with a window size of 128×128×128 and 50% overlap (aligned with common/official practice). To characterize performance beyond overlap similarity, we report three complementary metrics: Dice similarity coefficient (DSC; higher is better), false-negative volume (FNV; lower is better), and false-positive volume (FPV; lower is better). FNV directly quantifies missed tumor burden relative to the ground truth and is particularly informative for small/low-uptake lesions, whereas FPV constrains over-segmentation driven by physiological uptake or reconstruction noise.

Against nnU-Net and representative AutoPET Challenge methods, our approach achieves stable segmentation quality across both subsets and remains competitive in false-negative control (Table 1). A common trade-off observed among competing methods is that improvements in DSC may come at the expense of increased FPV, or reductions in FPV may increase FNV. Our results suggest that SDF-HOLO provides a more balanced operating point across DSC, FNV, and FPV. Notably on the UKT subset, we obtain a favorable combination of DSC and FNV, indicating improved sensitivity to subtle lesions without a marked increase in FPV.

Table 1 | Segmentation performance on AutoPET test subsets (Imu/UKT FDG). Higher DSC is better; lower FNV/FPV is better.

| Method | DSC | | FNV | | FPV | |
| --- | --- | --- | --- | --- | --- | --- |
| | Imu | UKT | Imu | UKT | Imu | UKT |
| nnU-Net [31] | 0.5017 | 0.8622 | 14.0836 | 1.6795 | 11.7733 | 6.6433 |
| Max.sh [32] | 0.4012 | 0.7483 | 30.2820 | 0.6570 | 38.8093 | 3.8532 |
| HKURad [33] | 0.6026 | 0.6973 | 10.8671 | 3.3258 | 4.8969 | 2.6552 |
| WukongRT [34] | 0.6204 | 0.7221 | 9.3683 | 1.1040 | 6.7212 | 3.5422 |
| BAMF AI | 0.6666 | 0.6973 | 11.3668 | 1.4420 | 3.1945 | 1.7017 |
| zero_sugar [35] | 0.6580 | 0.7373 | 12.3233 | 4.3788 | 0.5802 | 0.8209 |
| Shadab [36] | 0.6159 | 0.7270 | 11.4725 | 0.7998 | 2.1096 | 1.9246 |
| QuantIF [37] | 0.5800 | 0.7649 | 6.7282 | 2.2904 | 10.0465 | 0.7074 |
| Lennonlychan [38] | 0.6286 | 0.7454 | 12.5980 | 1.1815 | 1.8980 | 1.2379 |
| Airamatrix [39] | 0.6253 | 0.7556 | 12.5127 | 1.6105 | 1.8776 | 0.8604 |
| UIH_CRI_SIL [40] | 0.6194 | 0.7922 | 4.6745 | 0.9560 | 1.9500 | 2.2770 |
| StockholmTrio [41] | 0.5771 | 0.7684 | 8.4921 | 1.1839 | 1.9526 | 1.7990 |
| HussainAlasmawi | 0.6370 | 0.7802 | 2.4542 | 0.6365 | 3.3616 | 2.8756 |
| IKIM [42] | 0.6043 | 0.7938 | 2.7249 | 1.0422 | 4.9373 | 2.6586 |
| LesionTracer [43] | 0.6619 | 0.7702 | 1.7810 | 0.4310 | 1.7030 | 3.0189 |
| **Ours** | **0.6328** | **0.6933** | **2.2745** | **0.6447** | **3.0544** | **3.3270** |

Importantly, Imu and UKT correspond to different centers with distinct protocols and lesion distributions. Maintaining competitive performance on both subsets therefore better reflects robustness under cross-domain shift. We hypothesize that multimodal pretraining in SDF-HOLO stabilizes feature alignment across centers and helps preserve recall under heterogeneous acquisition conditions. Beyond quantitative metrics, we will provide case-level visual comparisons (Fig. 2) spanning large lesions, small lesions, and low-uptake lesions. Preliminary inspection suggests improved boundary consistency in weak-signal regions and reduced spurious expansions near physiological uptake patterns. To isolate the contributions of key design elements, we will further include ablations comparing CT-only, PET-only, naive concatenation, and dual-stream interaction, as well as variants with and without semantic anchoring, and report their impact on FNV and FPV.

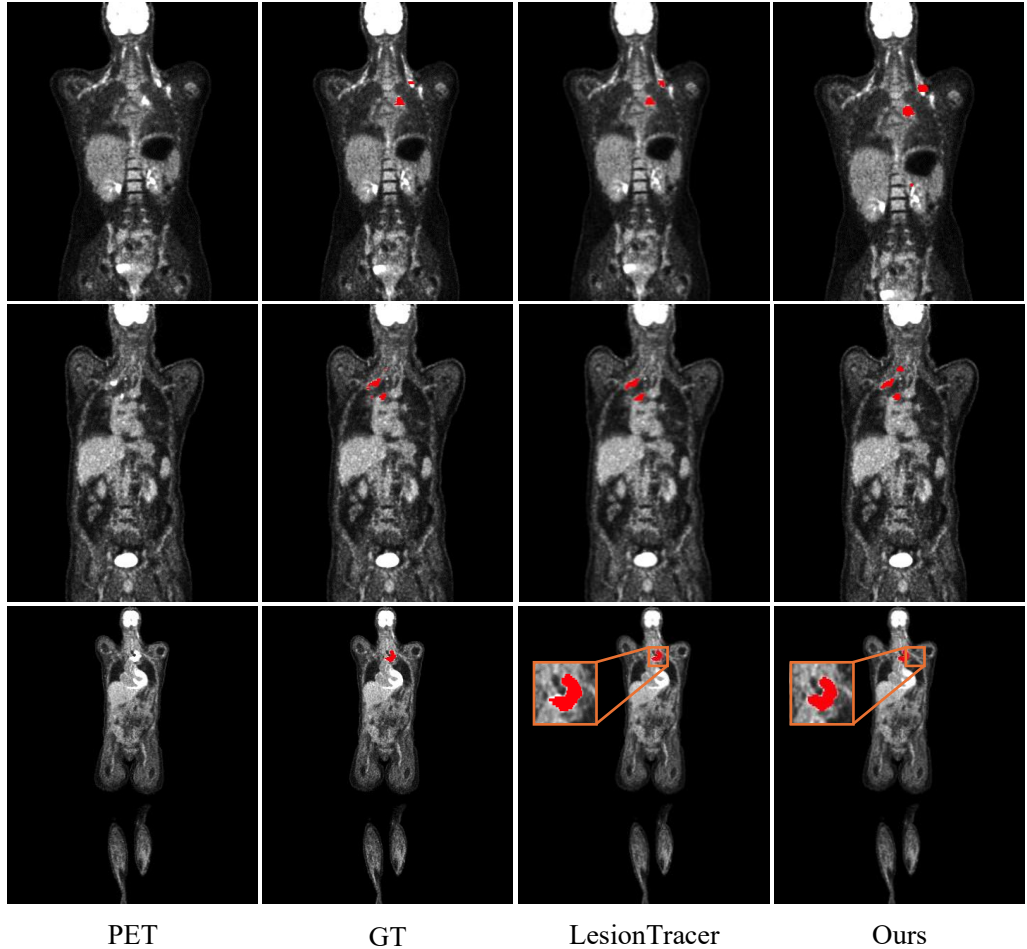

| PET | GT | LesionTracer | Ours |

Fig. 2 Showing the example cases of comparison between the ground truth lesion segmentation masks, the SOTA method-generated masks and the predicted lesion segmentation masks for lymphoma.

Overall, AutoPET results support the transferability of SDF-HOLO pretraining and its ability to maintain favorable false-negative control under domain shift and hard-case distributions, providing a reliable voxel-level foundation for subsequent system-level tasks including low-dose lesion detection, report generation, and metabolic network analysis.

4.3 Enhanced Report Generation

Automated radiology report generation is a stringent test of multimodal foundation models, requiring clinically coherent language that remains faithful to imaging evidence. In total-body PET/CT, the task is further complicated by heterogeneous anatomical–metabolic signals, ultra-long axial coverage, and structured radiological semantics that demand precise spatial grounding. We therefore evaluated report generation as a core capability of SDF-HOLO.

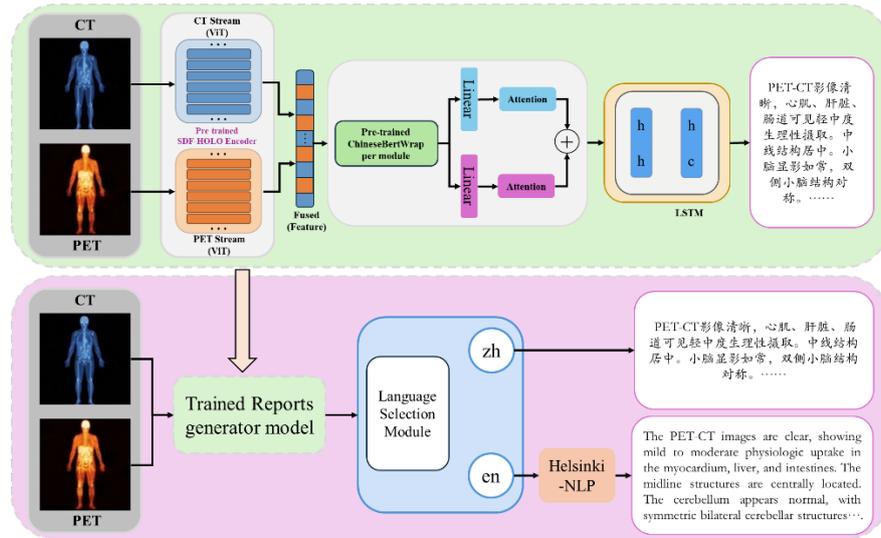

Fig.3 The architecture of the SDF-HOLO automated report generation and multi-lingual adaptation pipeline. The workflow operates in two stages: Top: The SDF-HOLO dual-stream encoders first extract and fuse anatomical (CT) and metabolic (PET) features. A semantic decoder, initialized with BERT and equipped with an LSTM-Attention mechanism, then generates the primary diagnostic report based on these fused representations. Bottom: To support international clinical collaboration, a multi-lingual adaptation module is integrated. It utilizes a keyword selection mechanism and the Helsinki-NLP translation model to convert the source report into, English while preserving medical semantic fidelity.

As illustrated in Fig.3, to address these challenges, we designed a streamlined generation pipeline. The visual features extracted by the SDF-HOLO dual-stream encoders are first projected into a semantic space and decoded into a structured report via a BERT-LSTM-based decoder. Crucially, to satisfy the need for global medical collaboration, we integrated a multi-lingual expansion module using Helsinki-NLP, enabling the system to output high-fidelity reports in both Chinese and English.

We first assessed report generation using standard metrics (BLEU-1/2/3/4, METEOR, ROUGE-L, CIDEr) and compared SDF-HOLO with competitive baselines including PET/CT report-generation pipelines and general medical vision–language models adapted for volumetric inputs. Across the internal test cohort, SDF-HOLO achieved consistent improvements over baselines on most metrics, with the largest gains observed for higher-order n-gram and sequence-level measures (e.g., BLEU-3/4 and ROUGE-L), indicating better long-range coherence and alignment with radiological reporting structure rather than reliance on generic templates. We further tested on an external multi-center cohort to probe robustness under heterogeneous scanners and reporting styles; despite an expected drop in absolute scores, SDF-HOLO maintained a stable margin over baselines, suggesting resilience to domain shift.

Table 2 Performance metrics of diagnostic report generation on test subsets

|  | Metric | | | | | | |
| --- | --- | --- | --- | --- | --- | --- | --- |
|  | BLEU_1 | BLEU_2 | BLEU_3 | BLEU_4 | METEOR | ROUGE | Cider |
| CNN | 0.1263 | 0.1137 | 0.1011 | 0.0884 | 0.1263 | 0.1263 | 0.0632 |
| Transformer | 0.1608 | 0.1448 | 0.1287 | 0.1126 | 0.1608 | 0.1608 | 0.0804 |
| **SDF-HOLO(ours)** | **0.4605** | **0.4408** | **0.4265** | **0.4148** | **0.4608** | **0.4608** | **0.3878** |

Because automatic metrics may underweight clinically critical errors, we analyzed three high-impact failure modes: hallucinated findings, incorrect anatomical localization, and omitted clinically relevant observations. Baseline models frequently produced metabolically active lesion statements without corresponding evidence or misattributed findings to incorrect organs, reflecting insufficient voxel-to-language grounding. In contrast, SDF-HOLO substantially reduced these errors and produced more anatomically plausible, evidence-consistent descriptions. These gains are consistent with SDF-HOLO's mask-guided semantic anchoring during pre-training, which enforces voxel–mask–text correspondence and improves fine-grained spatial–semantic alignment.

Total-body PET/CT reporting often requires integrating findings across multiple organs (e.g., primary tumor with nodal or distant metastases). SDF-HOLO generated more coherent multi-organ narratives, more reliably enumerating involved sites and preserving logical relations between primary and secondary findings. Compared with baselines, SDF-HOLO reduced fragmentation of multi-site descriptions and improved inclusion of clinically relevant negative findings, highlighting the value of whole-body context modeling for long-range dependency reasoning.

To assess clinical utility beyond text similarity, we performed blinded expert evaluation by nuclear medicine physicians/radiologists along accuracy, completeness, and readability. SDF-HOLO reports received higher scores than baselines, with the most pronounced improvements in accuracy and completeness for multi-lesion and complex metabolic cases. In a simulated reporting workflow, providing SDF-HOLO draft reports reduced report finalization time without compromising quality; junior readers benefited most in terms of completeness, while senior readers used the drafts as structured checklists to accelerate documentation.

Overall, SDF-HOLO's reporting gains can be attributed to three interacting design elements: dual-stream anatomical–metabolic encoding that preserves PET saliency under CT-dominant textures, whole-body context modeling that supports coherent total-body reasoning, and mask-guided semantic anchoring that strengthens voxel-to-language grounding and reduces hallucination. Together, these components move report generation beyond localized image captioning toward clinically faithful, system-level interpretation of total-body PET/CT.

4.4 Building a Healthy Metabolic Interaction Atlas

To extend SDF-HOLO's total-body representations beyond lesion-centric interpretation toward system-level physiology, we constructed an age-stratified reference atlas of metabolic interactions in a healthy cohort. The atlas characterizes coordinated and decoupled organ-to-organ patterns across the lifespan and provides a quantitative baseline for measuring disease-associated deviations at both organ and system scales.

We retrospectively screened healthy individuals from the overall dataset, totaling 181 cases, aged 12–82 years, stratified into a young group (12–45 years, n=30), a middle-aged group (46–65 years, n=105), and an older group (66–82 years, n=46). For selected analyses, the middle-aged group can be further divided at 55 years into lower- and higher-middle subgroups. For each subject, we extracted complementary features: (i) conventional interpretable metabolic intensity using organ-level SUVmean, and (ii) a 16-dimensional organ metabolic intensity embedding produced by SDF-HOLO to capture finer-grained metabolic "phenotypes" beyond a single scalar.

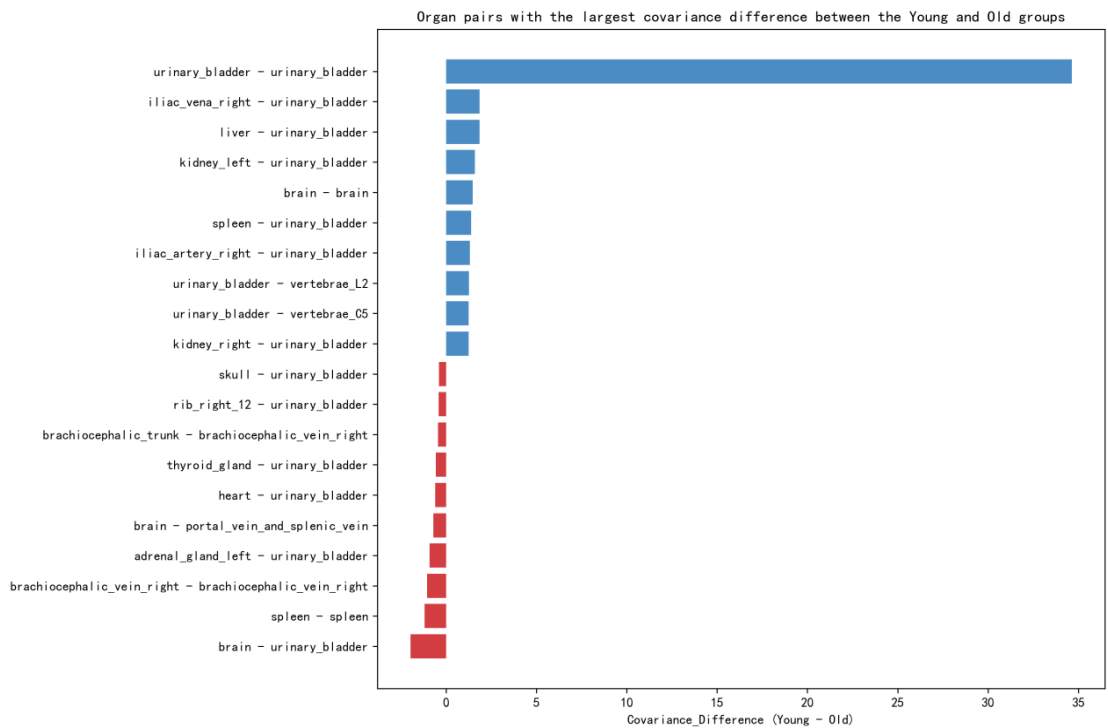

Fig. 4 Differences in organ-level covariance between the Young and Old groups. TOP 20 organ pairs with the largest covariance difference between the Young and Old groups including all 180 organs.

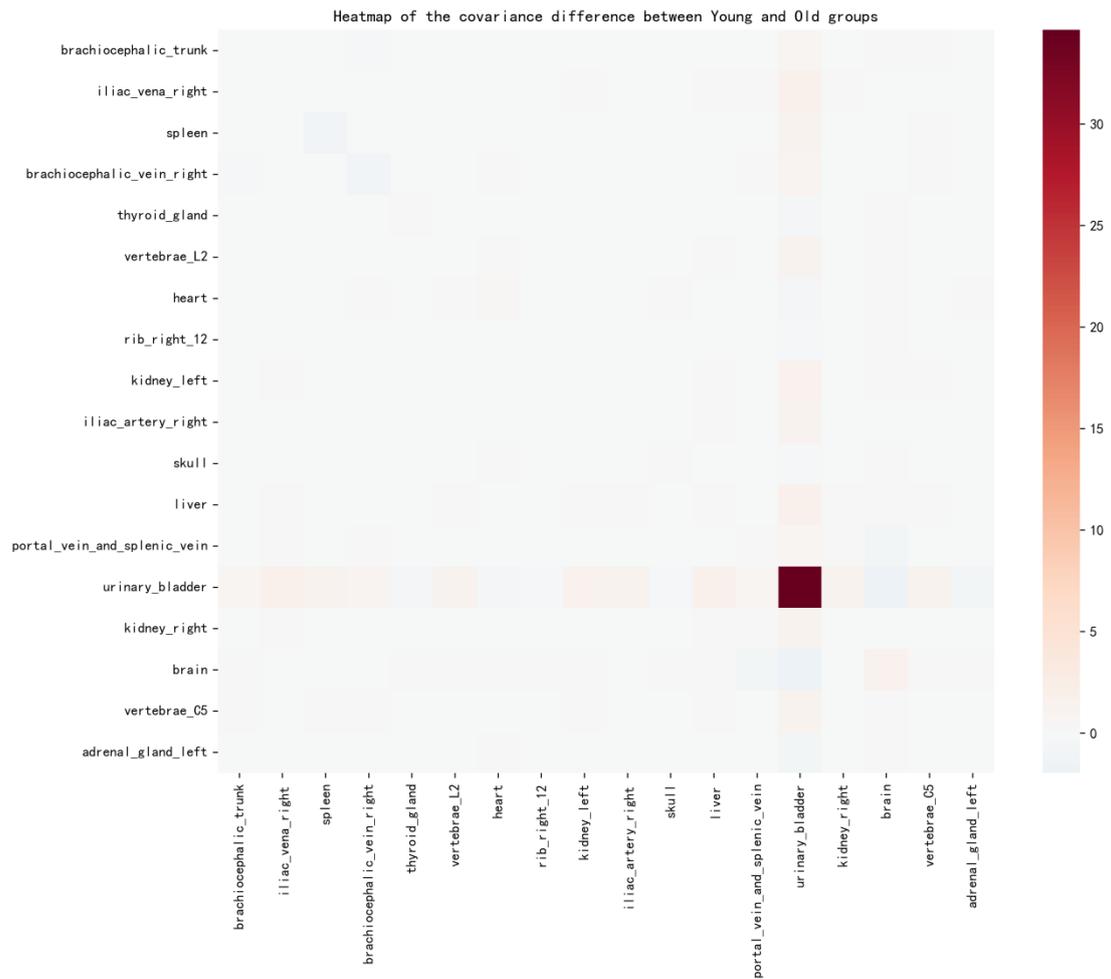

Fig. 5 Differences in organ-level covariance between the Young and Old groups. Heatmap of the covariance difference between Young and Old groups including all 180 organs.

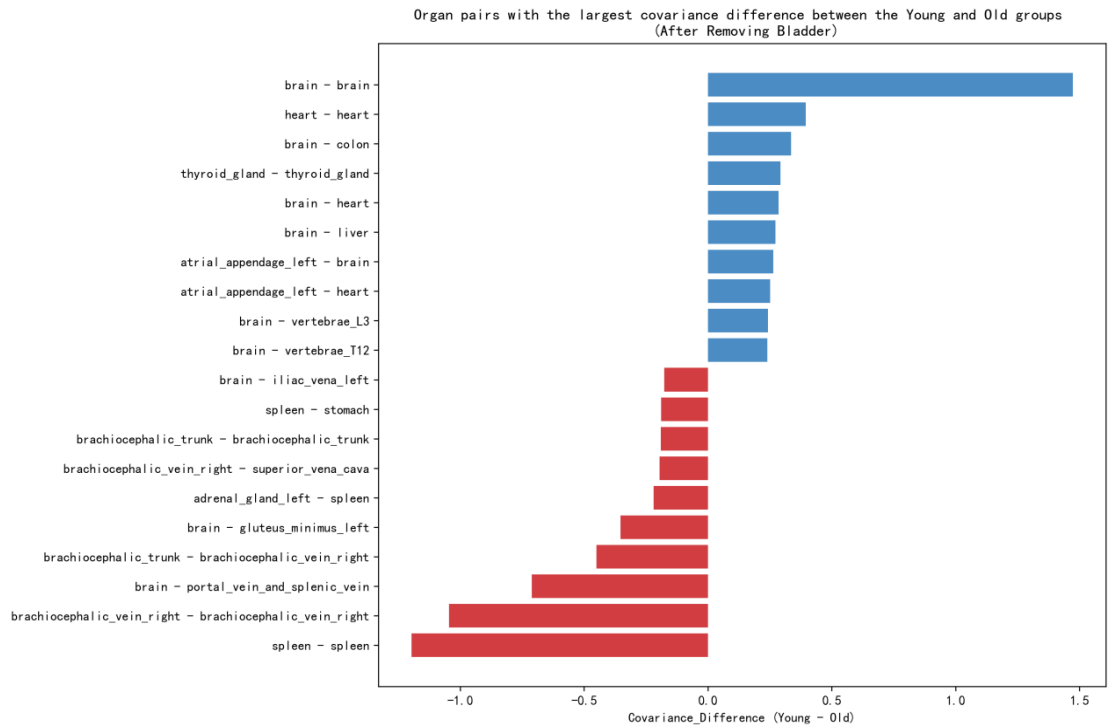

Fig. 6 Differences in organ-level covariance between the Young and Old groups. TOP 20 organ pairs with the largest covariance difference between the Young and Old groups, analysed after excluding the bladder, comprising 179 organs.

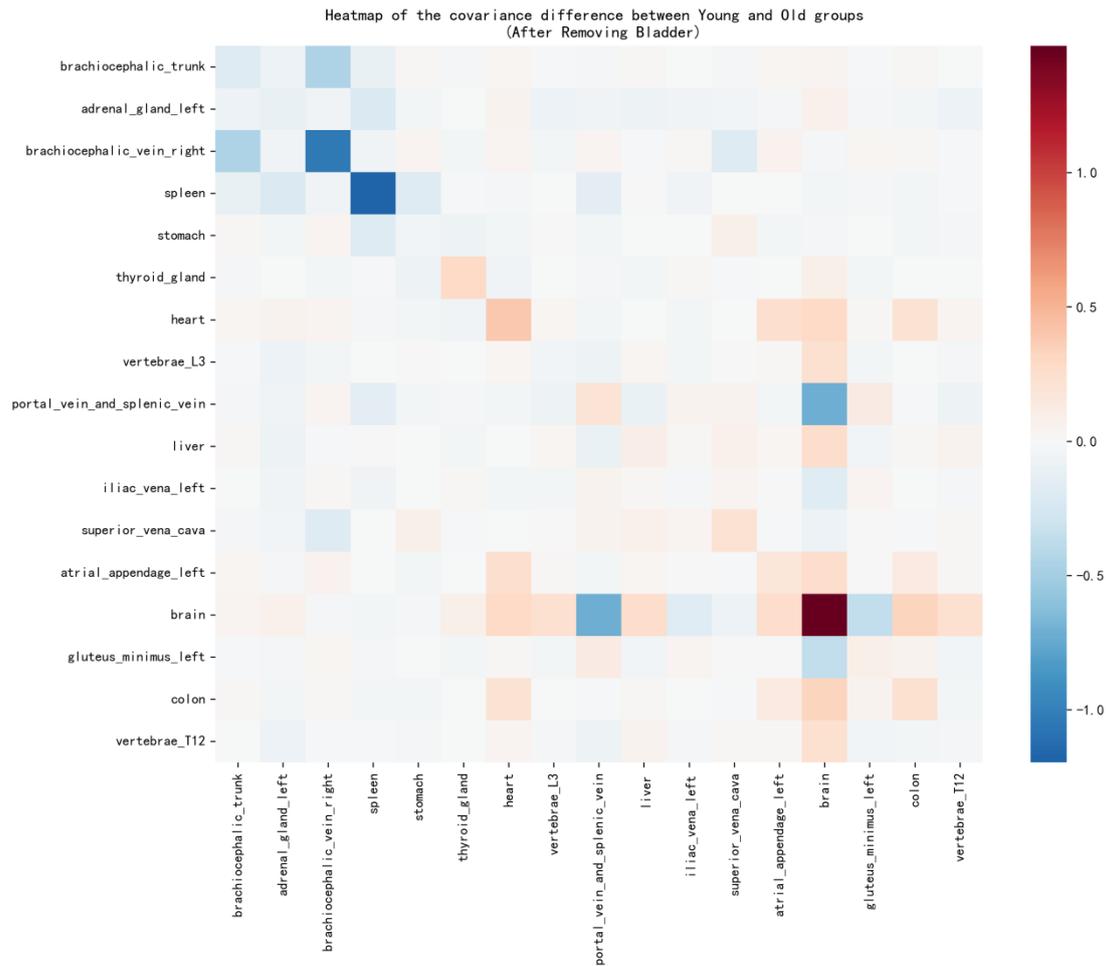

Fig. 7 Differences in organ-level covariance between the Young and Old groups. Heatmap of the covariance difference between Young and Old groups, analysed after excluding the bladder, comprising 179 organs.

Because FDG accumulates strongly in the urinary bladder approximately 1 hour post-injection, bladder uptake can dominate covariance structure and obscure subtler organ-to-organ relationships. We therefore quantified this effect and observed that including the bladder visibly amplifies high-variance blocks in covariance difference maps, masking broader interaction patterns. After excluding the bladder (179 organs), the covariance structure became more stable and revealed physiologically meaningful remodeling (Fig.4 -Fig.7). Accordingly, the atlas analyses below are performed in the bladder-excluded setting by default.

At the organ level, covariance difference analysis revealed prominent age-associated remodeling driven by core metabolic organs. The most marked positive difference between young and older cohorts was brain-related: the brain's self-covariance was higher in the young cohort by 1.47—a finding consistent with previous reports [44]. In contrast, the spleen exhibited the largest negative difference, with self-covariance lower in the young cohort by 1.20, whereas most other organs showed differences below 0.4. This phenomenon of "relatively preserved or even superior metabolic homeostasis in the spleen during aging compared to other degenerating organs" has been confirmed in animal studies [45].Together, these results indicate systematic reconfiguration of whole-body metabolic coupling during healthy aging, with organ-specific signatures that can be quantified in a unified framework.

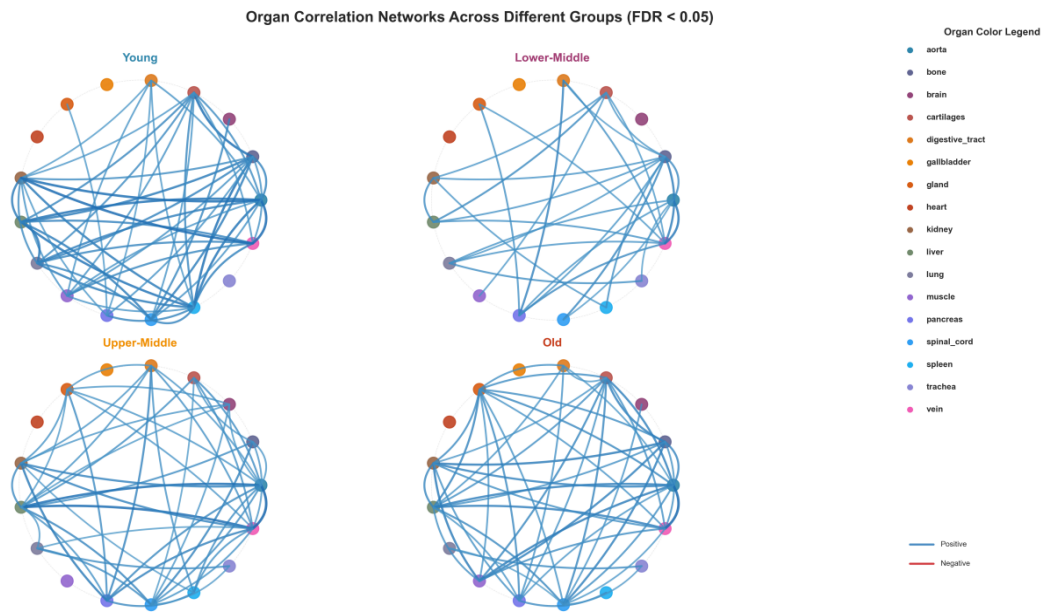

Fig. 8 Metabolic network structure diagrams for the Young, Middle (Lower-Middle and Upper-Middle), and Old groups (|r| ≥ 0.5 and FDR < 0.05). Nodes in the diagram represent representative key organs of the human body. The connecting lines (arcs) between nodes indicate the metabolic correlation (correlation coefficient) between two organs within the corresponding age group. Positive correlations are represented by blue arcs, and negative correlations by red arcs. The thickness and color intensity of an arc reflect the strength of the correlation.

We next constructed organ–organ interaction networks using thresholds |r| ≥ 0.5 with FDR < 0.05 and compared node connectivity across age strata. Vascular and structural hubs—including the aorta/arterial structures, venous structures, bone tissues, and the liver—maintained high connectivity across groups, suggesting a scaffold-like role in metabolic homeostasis. The circulatory system remains relatively stable throughout the aging process. As the core system sustaining life, its metabolic stability underscores the robust adaptability of the cardiovascular system to aging, with the slight increase observed in older age potentially linked to pathological factors such as arterial sclerosis [46, 47]. In parallel, aging was associated with progressive decoupling and reorganization of metabolic coordination, with the spleen and glandular organs occupying influential positions in the rewiring process (Fig. 8). In terms of major sources of variation, the young cohort was predominantly driven by digestive and endocrine-associated organs, the middle-aged cohort showed stronger contributions from reproductive-system indicators, and the older cohort displayed increased prominence of immune and circulatory components.

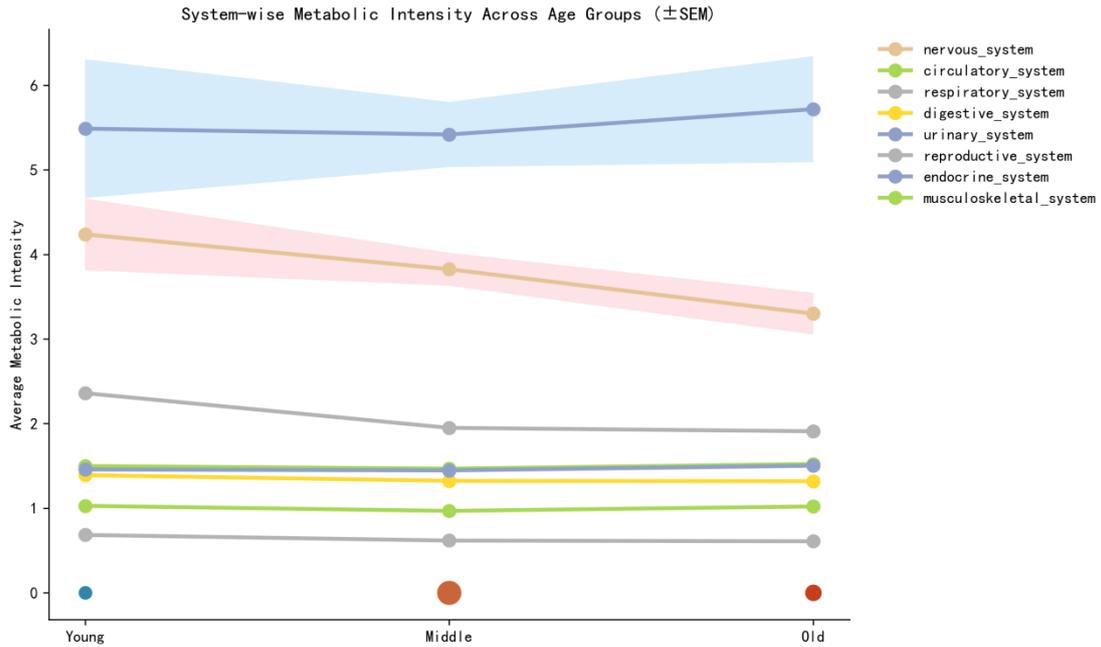

Fig. 9 Metabolic intensity differences and trends across eight systems in the three age groups. The eight major human body systems include: nervous system, circulatory system, respiratory system, digestive system, urinary system, reproductive system, endocrine system and musculoskeletal system. The colors for the age groups (Young/Middle/Old) are blue, green, and red, respectively.

Aggregating 180 organs into 8 physiological systems, we observed consistent system-level trends across age groups (Fig. 9). Neural, respiratory, digestive, and reproductive systems exhibited progressive declines in metabolic intensity with aging. Notably, the neural system decreased by 22.1% from young to older cohorts, with an accelerated decline in later life (13.7% drop from middle-aged to older). The reproductive system showed the largest early decline (17.4% from young to middle-aged) followed by a slower rate thereafter. Respiratory and digestive systems decreased by 10.9% and 5.2%, respectively. These trends align with clinical expectations while providing a quantitative, age-stratified baseline for defining 'healthy' metabolic ranges at the system level.

Integrating organ-level covariance remodeling, network connectivity, and system-level trajectories, we establish a Healthy Metabolic Interaction Atlas that supports downstream disease studies. Given a new subject, deviations from age-matched healthy baselines can be quantified across organs and systems, enabling interpretable characterization of tumor-associated systemic metabolic fingerprints and cross-organ abnormality localization

5. Discussion

5.1 Dual-stream anatomical–metabolic encoding for total-body PET/CT

Total-body PET/CT presents a distinct multimodal learning setting in which CT and PET encode complementary but highly heterogeneous signals. CT provides high-resolution structural context, whereas PET is sparse, high-dynamic-range, and sensitive to acquisition and physiological factors. Simple early-fusion strategies (e.g., channel stacking) can bias optimization toward CT texture statistics, potentially diminishing weak but clinically relevant PET signals, especially in low-uptake lesions or small tumor foci.

SDF-HOLO addresses this imbalance by learning anatomical and metabolic representations with modality-specific encoders and introducing cross-modal interaction at deeper layers. This design enables CT-derived spatial context to support PET interpretation while allowing PET-derived saliency to inform attention over subtle structural abnormalities. In downstream evaluation, this separation-then-interaction approach is consistent with improved robustness under protocol and site heterogeneity, and it provides a principled alternative to shallow fusion when modeling system-wide PET/CT volumes.

5.2 Mask-guided semantic anchoring for evidence-grounded language outputs

Radiology report generation from whole-body imaging requires both global clinical coherence and precise anatomical attribution. In total-body PET/CT, the semantic space is especially complex because a single scan spans multiple organ systems and numerous anatomical subregions, increasing the risk that models generate plausible but incorrectly localized statements. This is not merely a language modeling issue; it reflects insufficient constraints linking textual entities (organs, segments, laterality) to spatial evidence.

To improve spatial–semantic correspondence, SDF-HOLO introduces segmentation masks as explicit semantic anchors during pre-training and uses them to enforce voxel–mask–text alignment. This strategy can reduce errors in anatomical localization and mitigate unsupported statements by strengthening the model's ability to associate generated findings with anatomically defined regions. Importantly, while anchoring can improve evidence consistency, it does not eliminate the need for clinical oversight; remaining failure modes should be characterized with targeted evaluations that emphasize clinically meaningful errors (e.g., incorrect organ attribution, omissions of high-impact findings) rather than relying solely on n-gram similarity metrics.

5.3 System-wide metabolic interaction analysis enabled by foundation representations

Beyond lesion-centric tasks, total-body PET/CT creates an opportunity to quantify physiology at the scale of organ systems. Using organ-level metabolic features derived from the model, we constructed age-stratified reference patterns of metabolic covariance and network connectivity in healthy participants, and we further examined disease-associated deviations across cancer cohorts. These analyses suggest that model-derived representations may help summarize coordinated metabolic variation across organs and identify reproducible interaction patterns at the system level.

Nevertheless, network-level associations should be interpreted cautiously. Metabolic correlations in PET are influenced by tracer kinetics, acquisition timing, reconstruction settings, patient preparation (e.g., glucose level), and comorbidities; therefore, observed interaction patterns should not be over-interpreted as causal pathways. Future work should incorporate confounder-aware modeling and external validation to determine which network features are stable across sites and clinically actionable (e.g., prognostic stratification, treatment-response monitoring) versus those that reflect cohort- or protocol-specific factors.

5.4 Clinical translation and potential workflow integration

The clinical value of total-body PET/CT is coupled to the practical challenge of interpreting large, heterogeneous volumetric data under time constraints. A generalist model that supports multiple downstream tasks—such as lesion segmentation, low-dose lesion detection, and structured report generation—could contribute to workflow efficiency by standardizing outputs and providing consistent candidate findings for review. In particular, the combination of (i) multimodal representation learning tailored to PET/CT heterogeneity, (ii) whole-body context modeling, and (iii) evidence-grounded language generation offers a path toward assistive systems that prioritize safety and interpretability.

For deployment, however, several considerations are essential: prospective reader studies to quantify time savings and error profiles; calibration and uncertainty reporting for low-uptake or atypical findings; and integration with clinical reporting standards and institutional QA processes. Multilingual output can support cross-site collaboration, but translation fidelity for domain-specific terminology and institution-specific reporting templates should be assessed independently from core diagnostic accuracy.

5.5 Limitations and Future Directions

This study has several limitations. First, the training data are retrospective and, despite multi-center collection, may embed biases related to scanner types, reconstruction settings, and regional population characteristics. Prospective evaluation across broader geographic and demographic distributions will be required to establish generalizability and clinical safety. Second, segmentation masks used for semantic anchoring are produced by automated tools and label harmonization; residual errors in these masks could propagate into the learned alignment and downstream generation, motivating systematic quality assessment and robustness analyses to mask noise. Third, current system-level analyses largely rely on static uptake-derived summaries; dynamic imaging and kinetic modeling could provide more specific physiological characterization and reduce confounding from acquisition timing or perfusion effects.

Several directions may further strengthen the framework: (i) prospective, multi-site validation with standardized protocols and reader-centered endpoints; (ii) extension to multiple tracers and heterogeneous clinical indications beyond FDG-dominant oncology cohorts; (iii) incorporation of dynamic PET or time-resolved features to disentangle perfusion and metabolism; and (iv) multimodal integration with clinical variables and molecular assays to test whether imaging-derived system signatures add predictive value beyond established biomarkers. Together, these steps can clarify the conditions under which foundation representations for total-body PET/CT translate into reliable clinical benefit.

## 6. Conclusion

In summary, we curated a large multi-center total-body PET/CT dataset with paired radiology reports and developed SDF-HOLO, a multimodal generalist foundation model tailored to whole-body PET/CT analysis. Our framework integrates (i) an anatomical–metabolic dual-stream encoder with cross-modal interaction, (ii) hierarchical whole-body context modeling for long-axial reasoning, and (iii) mask-guided semantic anchoring that strengthens voxel-to-language grounding. Across representative downstream tasks—including automated tumor segmentation under domain shift, low-dose lesion detection, and bilingual report generation—SDF-HOLO demonstrates consistent performance gains or competitive trade-offs relative to strong baselines, while reducing clinically important failure modes such as localization errors and unsupported statements. Beyond focal lesion assessment, we show that model-derived organ-level features can support system-wide metabolic profiling and enable the construction of reference interaction patterns in healthy subjects, providing a quantitative baseline for studying disease-associated deviations. These results suggest that PET/CT-specific foundation representations can unify voxel-level tasks and higher-level phenotyping within a single computational framework. Future work should prioritize prospective multi-site validation, broader tracer and population coverage, and confounder-aware modeling to clarify generalizability and clinical utility in real-world deployment.